\documentclass[letterpaper, 10 pt, conference]{ieeeconf}  
\pdfoutput=1

\IEEEoverridecommandlockouts                              

\usepackage{graphicx}
\graphicspath{{}}

\usepackage{amsmath}
\usepackage{amssymb}
\usepackage[noadjust]{cite}

\usepackage{color}

\DeclareMathOperator*{\argmaxA}{arg\,max} 
\DeclareMathOperator{\Lagr}{\mathcal{L}}

\title{\LARGE \bf
Self-Supervised Surgical Tool Segmentation using Kinematic Information
}

\author{Cristian da Costa Rocha, Nicolas Padoy, and Benoit Rosa
\thanks{\textbf{Preprint accepted for the IEEE ICRA 2019 conference. Final typeset version will be available on https://ieeexplore.ieee.org}}
\thanks{*This work was supported by the French National Research Agency (ANR-18-CE19-0012-01, MACROS project) and by French state funds managed within the Investissements d'Avenir program by the ANR under reference ANR-11-LABX-0004 (LABEX CAMI).}
\thanks{Authors are with ICube, University of Strasbourg, CNRS, IHU Strasbourg, France. %
        {\tt\small b.rosa@unistra.fr}}%
}

\begin{document}

\maketitle
\thispagestyle{empty}
\pagestyle{empty}

\begin{abstract}

Surgical tool segmentation in endoscopic images is the first step towards pose estimation and (sub-)task automation in challenging minimally invasive surgical operations. While many approaches in the literature have shown great results using modern machine learning methods such as convolutional neural networks, the main bottleneck lies in the acquisition of a large number of manually-annotated images for efficient learning. This is especially true in surgical context, where patient-to-patient differences impede the overall generalizability. In order to cope with this lack of annotated data, we propose a self-supervised approach in a robot-assisted context. To our knowledge, the proposed approach is the first to make use of the kinematic model of the robot in order to generate training labels. The core contribution of the paper is to propose an optimization method to obtain good labels for training despite an unknown hand-eye calibration and an imprecise kinematic model. The labels can subsequently be  used for fine-tuning a fully-convolutional neural network for pixel-wise classification. As a result, the tool can be segmented in the endoscopic images without needing a single manually-annotated image. Experimental results on phantom and \textit{in vivo} datasets obtained using a flexible robotized endoscopy system are very promising. 

\end{abstract}

\section{Introduction}

Robot-assisted surgery is becoming a \textit{de facto} standard for many minimally invasive surgical operations, thanks to the unique ability to provide dexterity at the distal tip of miniature instruments. In order to access the most challenging pieces of anatomy, flexible instruments using continuum robot arms have been developed~\cite{burgner2015continuum}. Those instruments are typically more complex to model and control than classical rigidly linked instruments. For this reason, many methods for shape estimation have been developed using embedded sensors such as Fiber Bragg Gratings or Electromagnetic trackers~\cite{kim2014optimizing,shi2017shape}. Other promising approaches include vision-based approaches, which are appealing because in all endoscopic applications, a camera is already included in the system. In this paper, we concentrate on the first part of the pose estimation problem, i.e. the segmentation of the tools in endoscopic images. 

The main goal of segmentation is to provide a pixel-wise classification in order to determine the class that each pixel belongs to. In binary segmentation, each pixel will be either background or foreground, while the number of classes may be higher for a multi-class problem \cite{Pakhomov2017-qt}. Surgical tool segmentation is a challenging task and different approaches have been implemented in order to reach better results. Different strategies have been used based on markers to easily identify the surgical tool~\cite{Cabras2017-ub,Bouarfa2012-kk,wesierski2018instrument}, on handcrafted feature extraction~\cite{Allan2013-bt,Bouget2017-yp}, and on deep learning~\cite{Garcia-Peraza-Herrera2017-pu,Garcia-Peraza-Herrera2017-sa}. While being more appealing than marker-based approaches because they do not require modifying the tools, state-of-the-art marker-less methods are largely based on fully supervised learning~\cite{Allan2013-bt,Garcia-Peraza-Herrera2017-pu,Garcia-Peraza-Herrera2017-sa, Laina2017-mz}. For successful generalizability of supervised machine learning approaches training data should however be abundant. These data are typically acquired through manual annotation by an expert, which is both costly and time consuming. 

\begin{figure*}[!h]  
  \centering
  \includegraphics[width=0.7\textwidth]{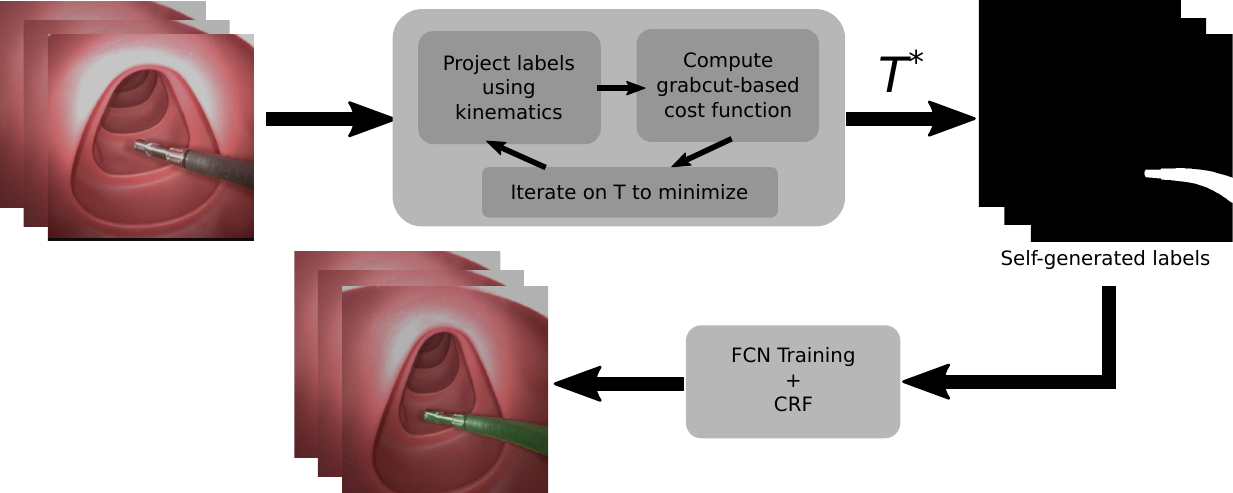}
  \caption[SSTS flow]{SSTS flow. Firstly, a transformation \( T^* \) between the camera and robot frame is found by the optimization of a Grabcut-based cost function on a given set of images. Next, the projection of the robot arm using \( T^* \) is obtained. These projections are used as labels for training an FCN, which can then be used together with CRF post-processing to perform tool segmentation inference. }
  \label{fig:ssts}
\end{figure*}

Unsupervised and self-supervised approaches have been introduced to tackle this issue. Indeed, a self-supervised scheme has no extra costs on the manufacturing of the instruments or need for time-consuming human annotation~\cite{Sermanet2017-cd}. On the other hand, the challenge lies on how to get reliable results without using ground truth information. Recently, Ross \textit{et al} proposed a generative adversarial network approach, which drastically reduces the number of necessary labeled data for surgical tool segmentation~\cite{ross2018exploiting}. In other domains, self-supervised approaches have been introduced for learning a robotic grasping task using large-scale automated data collection~\cite{levine2018learning}. 

In this paper, we present a self-supervised approach to surgical tool segmentation. In the context of robot-assisted surgery, using the kinematic model of the robot as a source of information is possible. To the best of our knowledge, this approach has however never been used for surgical tool segmentation due to various sources of errors (intra-corporeal environment, robot model, robot-tissue interaction, and hand-eye coordination). This is all the more true when using continuum robots, for which mechanical models can be very inaccurate, e.g. when external forces are applied upon contact with the anatomy. In order to cope with this problem, we propose a two-step algorithm (Fig.~\ref{fig:ssts}). Using a few randomly-selected images from the surgery together with the associated joint values, the first step will iteratively optimize the hand-eye calibration in order to generate good labels by projecting the robot model onto the image and maximizing a purposely developed cost function. In order to generalize the results to the rest of the surgery while relaxing the dependency to the robot kinematic model, we subsequently fine-tune a Fully Convolutional Neural network (FCN) using the labels obtained at the end of the first step. The developed approach can be applied to any surgical robotic system, continuum robot or not, provided that we have access to a kinematic model of the instrument, together with joint values synchronized with image acquisition. Validation was performed on several phantom and in vivo datasets acquired with a flexible robotic endoscopy system. Results show a very promising performance of the proposed approach, in the presence of potentially large mechanical model errors coming from firm contact with the anatomy.

\section{Self-Supervised Tool Segmentation}

\begin{figure*}[!ht]  
  \centering
  \includegraphics[width=\textwidth]{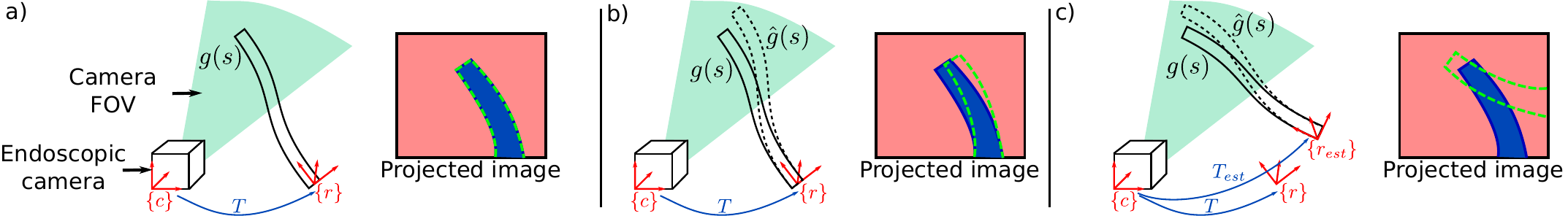}
  \caption[Different situations for the projected image]{Different situations for the projected image. In a) the projected image is represented by accurate T and shape g(q, s) information; b) gives the projected image with accurate T but inaccurate shape estimation \( \hat{g} \)(q, s) and c) shows the projected image with inaccurate T and shape estimation  \( \hat{g} \)(q, s). }
  \label{fig:scheme}
\end{figure*}


Let's consider the situation depicted on Fig.~\ref{fig:scheme}, where a continuum robot arm is inserted through a given access port into the body lumen or cavity (either a separated port in laparoscopy, or an instrument channel in endoscopy). The shape \( g(q, s) \) of the robot can be estimated using the kinematic model $g$, depending on the joint variables $q$ and parametrized by the arc-length $s$. This model is typically estimated in the robot base frame $\{r\}$. For simplicity, we assume here that the camera is calibrated --having \( k_{c} \) as the camera calibration matrix-- and that images are undistorted. If one knows the transformation $T$ between the camera frame $\{c\}$ and the robot frame $\{r\}$, as well as the shape and diameter of the robotic tool, the robot shape can be projected onto the image plane. The area delimited by the projection of the robot onto the image is denoted \( \Omega_{p}(q, T) \), and the set of image pixels that belong to it is noted \( y(q, T) \). Following a similar method as in~\cite{Pezzementi2009-ek}, we use a 3D rendering engine to project the whole robot shape onto the image. 

%


Rendering the robot's model on the image plane has, however, many limitations. First, kinematic models are often inaccurate, due to various nonlinearities and unmodeled phenomena (e.g. cable friction). This is shown in Fig.~\ref{fig:scheme}b, where the estimated model  $\hat{g}(q, s)$ does not match the actual instrument shape $g(q, s)$, leading to a slightly offset projection (dotted green) with respect to the instrument's actual position in the image (blue area). Moreover, without resorting to external sensors, the transformation $T$ is difficult to calibrate precisely beforehand. Using a bad estimate $T_{est}$ of the projection $T$ can lead to vary bad projections onto the image, as depicted on Fig.~\ref{fig:scheme}c. 

The proposed algorithm is named SSTS --Self-Supervised Tool Segmentation-- and its overall workflow is shown on Fig.~\ref{fig:ssts}. The core idea is to use the shape estimate coming from the kinematic model in an iterative optimization scheme, in order to find an estimate $T^*$ which minimizes the projection error onto the image (note that $T^*$ is not the optimal hand-eye calibration matrix due to robot modeling errors). Without manually-labeled ground truth this error is not trivial to estimate. We propose using a custom cost function defined from the results of a Grabcut-based tool segmentation initialized from the projected labels. In order to minimize the influence of the kinematic modeling errors, a few frames featuring the robot in different poses are acquired at the beginning of the surgery. After convergence, the obtained projections are used as labels for fine-tuning an FCN, which will subsequently be used for segmentation inference in the remaining of the surgery. These steps are detailed in the next two subsections. 

\subsection{Optimization of a Grabcut-based Cost Function}
\label{sec:optimization}

The goal of this optimization step is to find, given a set of images with their associated shape estimates coming from the kinematic model, a transformation $T^*$ which minimizes the error in the projection \( y(q, T) \) with respect to the actual position of the tool in the image. This is a complex task, given the uncertainties in the mechanical model and the fact that no ground truth is used for estimating the projection error. In order to evaluate the goodness of the fit between the robotic-projected labels and the image, we propose a cost function based on the Grabcut algorithm. The Grabcut algorithm \cite{Rother2004-tf} is a well-known segmentation method in Computer Vision. Given a few pixels in the image labelled as foreground and background, it segments the whole image using a graph-based algorithm. 

\begin{figure}[t]  
  \centering
  \includegraphics[width=\columnwidth]{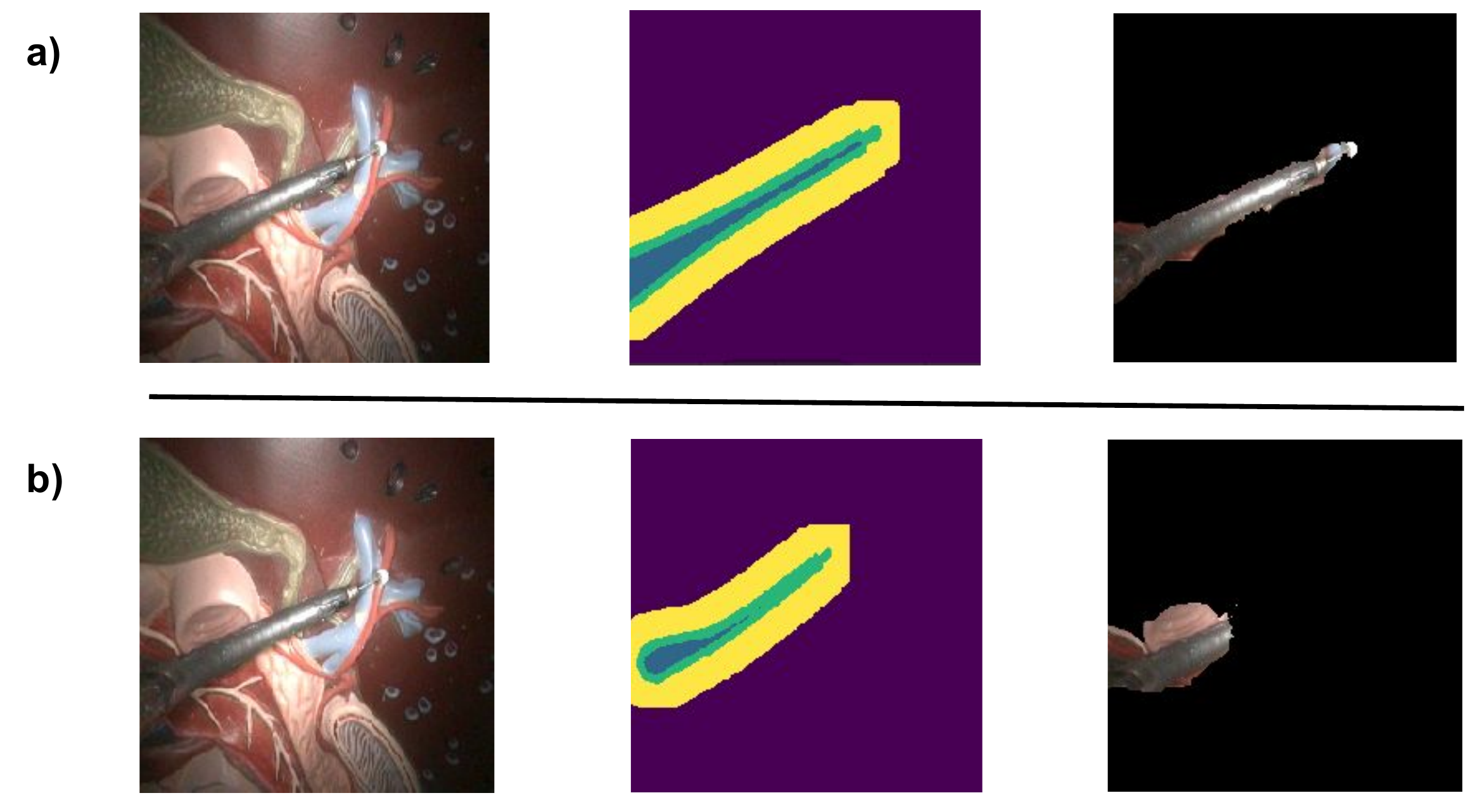}
  \caption[Grabcut segmentation given good and bad transformations]{Grabcut segmentation given good and bad transformations. With a) representing a good T and b) a bad T, it is possible to see the input image, the classes sure foreground (blue), likely foreground (green), likely background (yellow) and sure background (purple), and the Grabcut output.}
  \label{fig:grabcutseg}
\end{figure}

The intuition in our proposed approach is to consider that if the projected tool area $\Omega_{p}$ is accurate, giving this area as a set of foreground labels to the grabcut algorithm would yield a similar output in terms of foreground area. If, however, the transformation or the model are highly inaccurate, $\Omega_{p}$ will intersect both the tool and the background (or won't intersect the tool at all in the image), and the output of the grabcut algorithm will be very different from the $\Omega_{p}$ area (see Fig.~\ref{fig:grabcutseg}). In the following, we detail how this intuition can be turned into a cost function for optimizing the hand-eye calibration. 

Four classes are provided to the Grabcut algorithm: sure background, likely-background, sure foreground and likely-foreground. In order to account for model uncertainties, the projected area $\Omega_{p}$ is eroded to build the sure foreground area using a $7\times7$ kernel, while $\Omega_{p}$ is used as the likely-foreground area. Similarly, a $24\times24$ dilation is used to build the likely-background area, while the rest of the image is used to build the sure background area. The grabcut algorithm is then applied and the output labels are then used to compute a cost-function. 

Intuitively, if a good transformation T is given, the output of the Grabcut segmentation \textit{h} should have a high IoU (Intersection over Union) in comparison to the projected labels \( y(q, T) \) -- meaning that \textit{h} will have a high number of true positives and a low number of false positives and false negatives. In this context, the specificity (SPC) is also important and the pixels which are assumed to be likely background should not be segmented as foreground after the Grabcut segmentation. In these lines, even though the false negatives in the likely background area have an impact in the IoU, an ideal cost function would be one that could have a high penalization for false negatives while also evaluating the number of true positives and false positives. With that in mind, for evaluating a good transformation T, a simple harmonic mean between the IoU and the specificity, which is defined as \( F^\prime_1 \), is used as a cost function. For this evaluation, the pixels in the sure background area are not taken into account because the output of Grabcut will not consider this area for the segmentation : 
\begin{equation} 
 F^\prime_1(y(q, T), h) = 2 \frac{IoU(y(q, T), h)*SPC(y(q, T), h)}{IoU(y(q, T), h)+SPC(y(q, T), h)}.
\end{equation}

The optimization of the cost function is performed using a stochastic branch-and-bound algorithm which will explore the search space avoiding the convergence in local minima \cite{Papazov2011-hk,gruijthuijsen2016automatic}. The main goal is to find a transformation T which maximizes the \( F^\prime_1 \) score between the segmented output from the GrabCut algorithm and the projection of the mechanical model. The search is performed on the rigid transformation space SE3. This algorithm stochastically explores a tree splitting the search space into smaller spaces and evaluates the cost function at a randomly sampled point for each split. To summarize, the definition of the Grabcut-based SSTS optimisation step can be given by the formula below, where the main goal is to find a transformation \( T^* \) where \( F^\prime_1 \) is maximum :
\begin{equation} 
 T^* = \argmaxA_T F^\prime_1(y(q, T), h).
\end{equation}

\subsection{Fully Convolutional Network for Segmentation}
\label{sec:FCN}

Once the optimal transformation $T^*$ for a set of images has been found, one needs to generalize this result to the rest of the video. Using Grabcut with the projected labels \( y(q, T^*) \) could be a simple solution, but it has limitations. On some sample images, the projected labels may be far from the tool, especially when interacting with tissues, which may result in locally poor results such as the one displayed on Fig.~\ref{fig:grabcutseg}b.  

In order to tackle this issue, we propose to exploit the images used for the optimization of transformation $T$ and the resulting projected labels to train a machine learning model in order to perform the segmentation. This has several advantages: the labels are of better quality since they were optimized along with the transformation and the use of several images where the tool is in different positions helps reduce the effect of potential noisy labels.

A Fully Convolutional Network (FCN) was built for performing the segmentation. FCNs are commonly used in semantic segmentation algorithms, where it is also necessary to have a pixelwise prediction \cite{Long2015-wb}. The proposed FCN is based on ResNet18, pre-trained on Imagenet and fine-tuned on surgical tool presence detection on endoscopic images \cite{Vardazaryan2018-du}. In this case, just the first convolutional layer and the next two residual blocks of ResNet are used in the architecture, followed by two resize-convolutions with nearest-neighbour interpolation for upsampling the output to be the same size as the input. Finally, a 1 x 1 convolution layer is added to extract the score maps. As the training is meant to be online, at the beginning of each surgery, the depth of the network should not be very high, given that not many images are provided for training and also that the training is expected to be short, in the range of a few minutes. 

\begin{figure}[!ht]  
  \centering
  \includegraphics[width=\columnwidth]{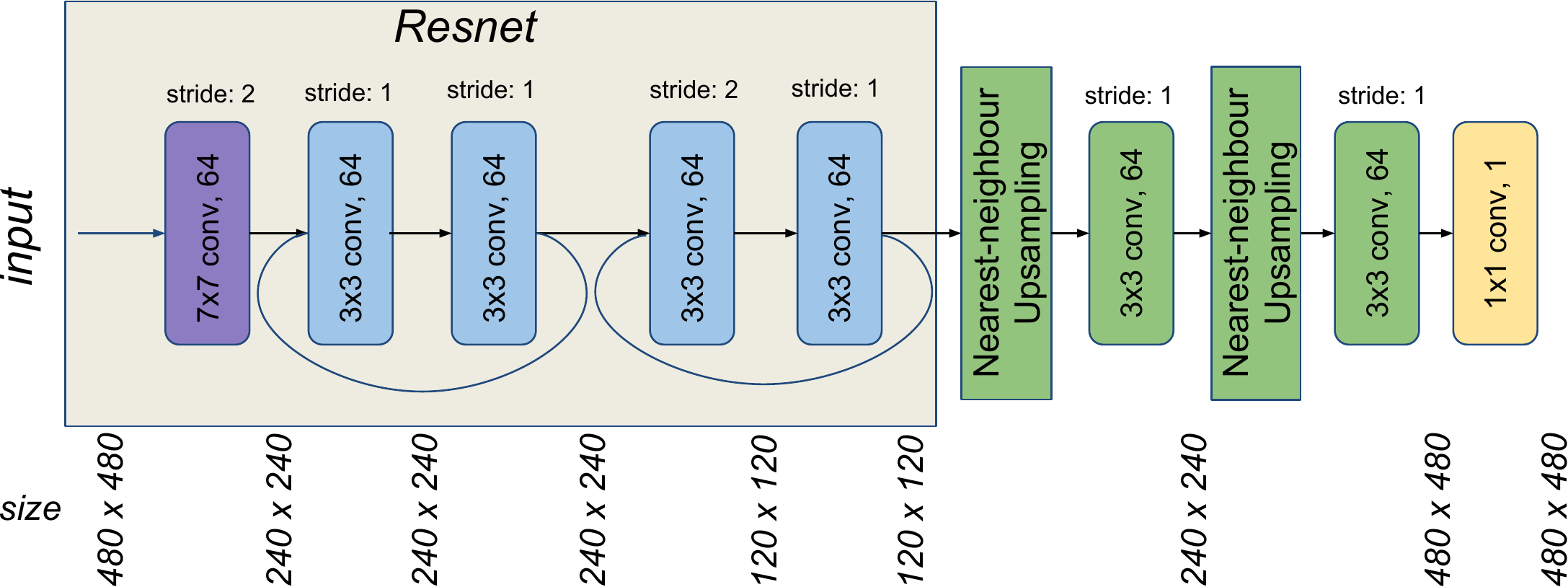}
  \caption[FCN architecture used for SSTS]{FCN architecture used for SSTS.}
  \label{fig:resnet}
\end{figure}

The network takes an RGB image with a size of $480 \times 480$ as input, which has its size reduced by half in the first convolutional layer of $7 \times 7$, with 64 filters. The first residual block has two convolutional layers with $3 \times 3$ kernels and stride of 1, with a skip connection between the two convolutional layers. In sequence, the second residual block also has two convolutional layers with $3 \times 3$ kernels with a skip connection between them, but in this case, the first convolutional layer has a stride of 2, which decreases by half the size of the output. In order to have the final score map with the same size as the input image the output of the second residual block is upsampled by using a resize function based on nearest-neighbour followed by a $3 \times 3$ convolutional layer~\cite{Odena2016-pp}. This upsampling strategy is applied twice, having a $1 \times 1$ convolution with a filter depth of 1 for producing the final score maps. The overall architecture of the FCN is shown on Fig.~\ref{fig:resnet}.

The FCN is trained using Adam optimization for minimizing a weighted cross-entropy function between the predicted pixels to the given foreground/background pixel labels. The loss function writes : 
\begin{equation} 
\begin{split}
\Lagr &= -\frac{1}{n} \sum_{i=0}^{n} \Big[ \delta \ y^i(q, T) \log ( \hat{y}^i )  \\ &+ (1 - y^i(q, T) ) \log (1 -  \hat{y}^i ) \Big] + \frac{\lambda }{2} \lVert w \rVert^2_2 . 
\end{split}
\end{equation} 

The projected labels are represented by \( y^i(q, T) \) at each pixel position $i$, and \( \hat{y}^i \) is the output of the score map after the application of a sigmoid function in order to represent foreground and background classes as probabilities. \( n \) represents the number of pixels in the batch of images and \( L_2 \) regularization is performed on all the network weights \( w \). The weight decay parameter is given by \( \lambda = 10^{-4} \) with the goal of avoiding overfitting and for reaching a higher degree of generalization. The momentum used is 0.9 with a learning rate fixed to \( lr = 0.01 \), where all layers have their weights updated after each epoch. The layers which are not from ResNet are randomly initialized at the start of the training. In order to give a higher penalization to false positives, the weight parameter \( \delta \) is fixed to 3.

In order to improve the generalizability of the results to other parts of the surgery, data augmentation is performed during training. For each epoch, the images in the batch have their hue, saturation and contrast slightly modified. The hue value is randomly modified with a maximum deviation of 1\%, whereas the saturation and contrast are randomly modified within a 20\% range. In addition, the images in the batch are also flipped upside down or from left to right, and randomly cropped.

Finally, one should note that the final segmentation output may be slightly noisy. False positive pixels may be appearing near the surgical tool, as well as sometimes in challenging areas featuring specular reflection and blur. Similarly to~\cite{Chen2018-gk,Papandreou2015-hn}, we post-process the output of the FCN using Conditional Random Fields~\cite{Krahenbuhl2012-gr} to produce more accurate segmentation results. 

\section{Experimental validation}
\label{sec:validation}

The experimental validation was performed using the STRAS robot~\cite{Donno2013-kx}, which is a teleoperated prototype for robotized flexible endoscopic surgery~\cite{Legner2017-xf}. It has an endoscopic camera and the possibility for positioning two instruments: left and right robot arms. The robot arms used in STRAS are flexible cable-actuated instruments, and therefore can be considered as continuum robots. Each arm has 3 joint angles/positions: the whole body rotation, the insertion of the tool, and the cable-actuated bending. Using the constant curvature assumption, those joint angles can be used to compute the forward kinematic model of the robot, following equations detailed in~\cite{Webster2010-kl}. As discussed earlier, this model does not consider non-linearities such as cable friction and can become very inaccurate as soon as external forces or wrenches are applied on the robot body. 

During the operation of the robot, the user sends commands through a user interface, which are used to servo the motors controlling the joints of the robot. The positions of the motors are recorded into an array of joint values $q$. Camera images are acquired through an acquisition board in a synchronized fashion, resulting in RGB images with a resolution of 570x760 pixels. The STRAS robot has a fixed focal camera, therefore it has only been calibrated once for all the experiments, using the standard calibration procedure from~\cite{zhang2000flexible}.

Three different datasets were used for validating the proposed approach. The first two were recorded on the benchtop, using respectively a plastic model of the human digestive system and a silicone model of the human colon. For both datasets, thereafter named \textit{phantom 1} and \textit{phantom 2}, movements of the environment were manually induced. The third dataset was acquired \textit{in vivo} during a surgery in a porcine model. It features phases of tissue interaction, dissection, smoke, partial and complete instrument occlusions. Table~\ref{table:datacollection} summarizes the different datasets' number of images, as well as ground truth (GT) images, which were randomly selected and manually segmented for validation purposes.

\begin{table}[h]
\caption{Number of images and GT in each dataset.}
\label{table:datacollection}
\begin{center}
\begin{tabular}{|c||c|c|}
\hline
\textbf{Dataset} &  \textbf{Number of images}  & \textbf{Number of GT images} \\ 
\hline\hline
\textit{phantom 1}  & 910    & 30  \\ \hline
\textit{phantom 2}  & 2737    & 30  \\ \hline
\textit{in vivo}  & 481   &  30  \\ \hline
\end{tabular}
\end{center}
\end{table}

The algorithms were implemented in Python, using Tensorflow for the analysis involving deep learning, VTK for the 3D volume rendering of the continuum robot (used to generate the projected labels), and OpenCV for the Grabcut segmentation and other image processing tasks. Results were generated using an Intel(R) Core(TM) i7-3930k (3.20GHz) with 32GB of RAM and a GeForce GTX 1080 Ti GPU.  

\section{Results}
\label{sec:results}

This section presents the results obtained with the proposed approach. For each dataset, 19 images were randomly selected in the beginning of the video and used for both finding an optimal transformation $T^*$ and fine-tuning an FCN for tool segmentation. One notable exception is the \textit{in vivo} dataset, for which the video corresponding to the beginning of the surgery was not available. For this dataset, images where the instrument is not in contact with the anatomy were selected. This is in agreement with the envisioned workflow, in which images from the beginning of the surgery are used, where the doctor mainly moves the instruments in free space before starting manipulating tissues. 

\subsection{Grabcut-based Branch-and-bound Optimization}

\begin{figure}[t]  
  \centering
  \includegraphics[width=0.75\columnwidth]{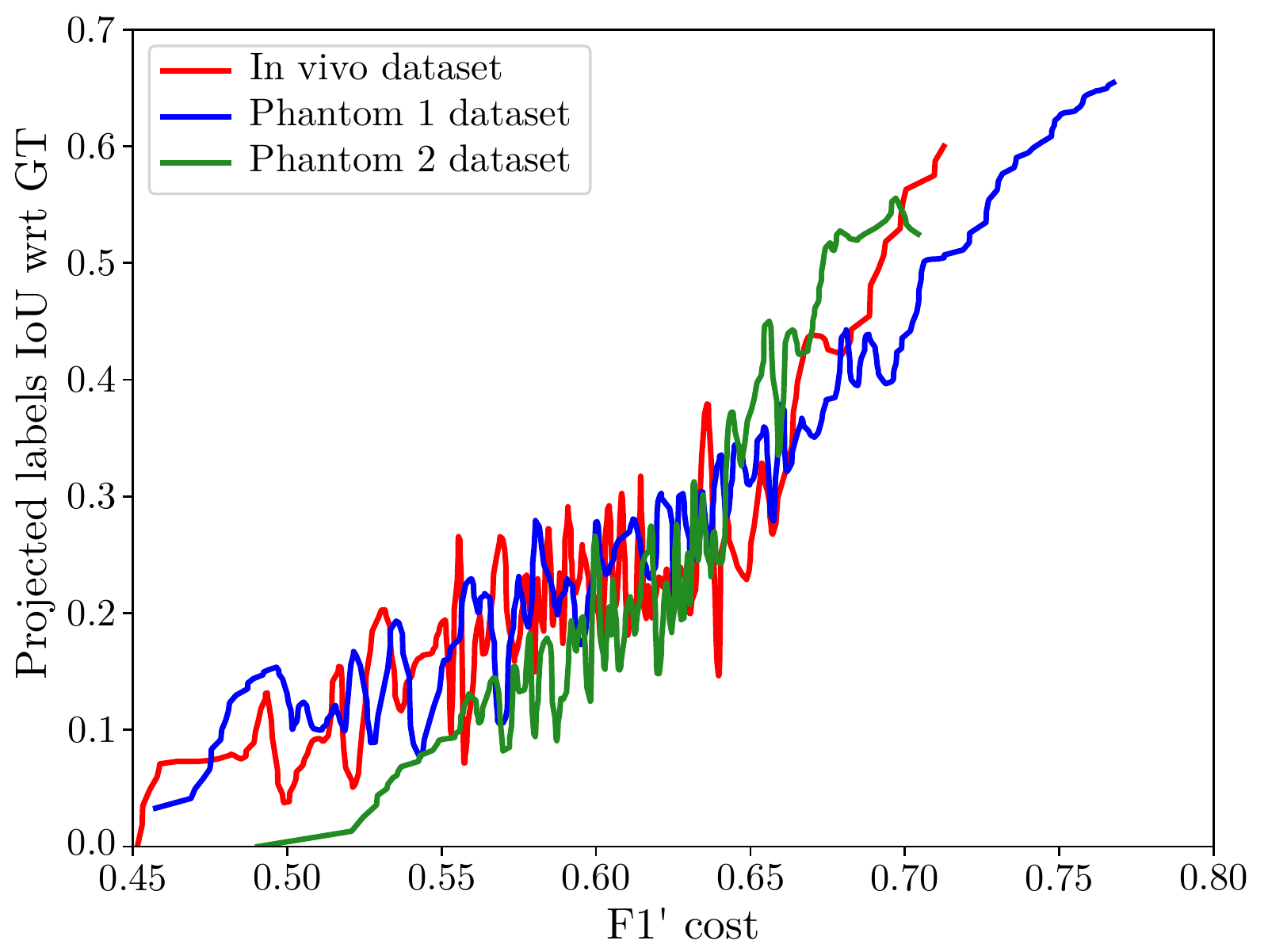}
  \caption[Cost function comparison.]{IoU between the self-generated labels and the GT as a function of \( F^\prime_1 \) during branch-and-bound optimization.}
  \label{fig:ssts_optim}
\end{figure}

Fig.~\ref{fig:ssts_optim} presents typical convergence results obtained, for each of the 3 datasets, by using 19 images during the SSTS optimization step. The Branch and Bound optimization was run for 300 iterations and the \( F^\prime_1 \) cost function as well as the self-generated labels were saved for each iteration. The Intersection over Union (IoU) between those labels and manually segmented GT was computed and plotted on the y-axis. The higher this IoU, the more accurate the forthcoming FCN training step will be. While the obtained IoU never reaches 1 due to model inaccuracies, it is however almost monotonically related to the cost function \( F^\prime_1 \). Indeed, linear correlation coefficients are respectively of 0.82, 0.74 and 0.6 for the phantom 1, phantom 2 and in vivo datasets (the lower correlation for the in vivo dataset being explained by a clear change of slope around \( F^\prime_1 = 0.65 \). These results validate the fact that maximizing the cost function in the SSTS optimization step also maximizes the IoU with respect to the GT (but without GT in the optimization).

\subsection{FCN fine-tuning}

Following SSTS optimization, the FCN was fine-tuned using the procedure described above. For validation purposes, the selected images were also manually annotated in order to train the same FCN in a fully supervised way. For the testing part, 11 manually annotated images were used for generating common metrics such as Precision, Recall, IoU. The overall time was on average 2600 seconds (900 for the optimization and 1700 for the FCN training with 1000 epochs). 

\begin{table}[h]
\caption{SSTS results comparison with fully supervised learning (FSL) and Grabcut.}
\centering
\begin{tabular}{|c||c|c|c|c|}

\hline
\textbf{Approach} &  \textbf{Accuracy}  & \textbf{IoU}  & \textbf{Recall}  & \textbf{Precision}  \\ \hline\hline

\multicolumn{5}{|c|}{\textbf{Phantom 1}} \\
\hline
SSTS  & 0.99    & 0.86  & 0.90  &  0.92    \\ \hline
FSL  & 0.99   &  0.87 &   0.92   &   0.93  \\ \hline
Grabcut  & 0.97    & 0.56  & 0.86  &  0.61    \\ \hline
\hline
\multicolumn{5}{|c|}{\textbf{Phantom 2}} \\
\hline
SSTS & 0.98    & 0.78  & 0.88  &  0.87    \\ \hline
FSL  & 0.98  &  0.84   &   0.88   &   0.94 \\ \hline
Grabcut & 0.95    & 0.49  & 0.66  &  0.66    \\ \hline
\hline
\multicolumn{5}{|c|}{\textbf{In Vivo}} \\
\hline
SSTS  & 0.97    & 0.62  & 0.66  &  0.91    \\ \hline
FSL  &  0.98  &  0.72   &   0.73    &  0.98   \\ \hline
Grabcut  & 0.96    & 0.55  & 0.73  &  0.69    \\ \hline

\end{tabular}
\label{table:sstsresults}
\end{table}

\begin{figure}[!ht]  
  \centering
  \includegraphics[width=0.8\columnwidth, height = 4cm]{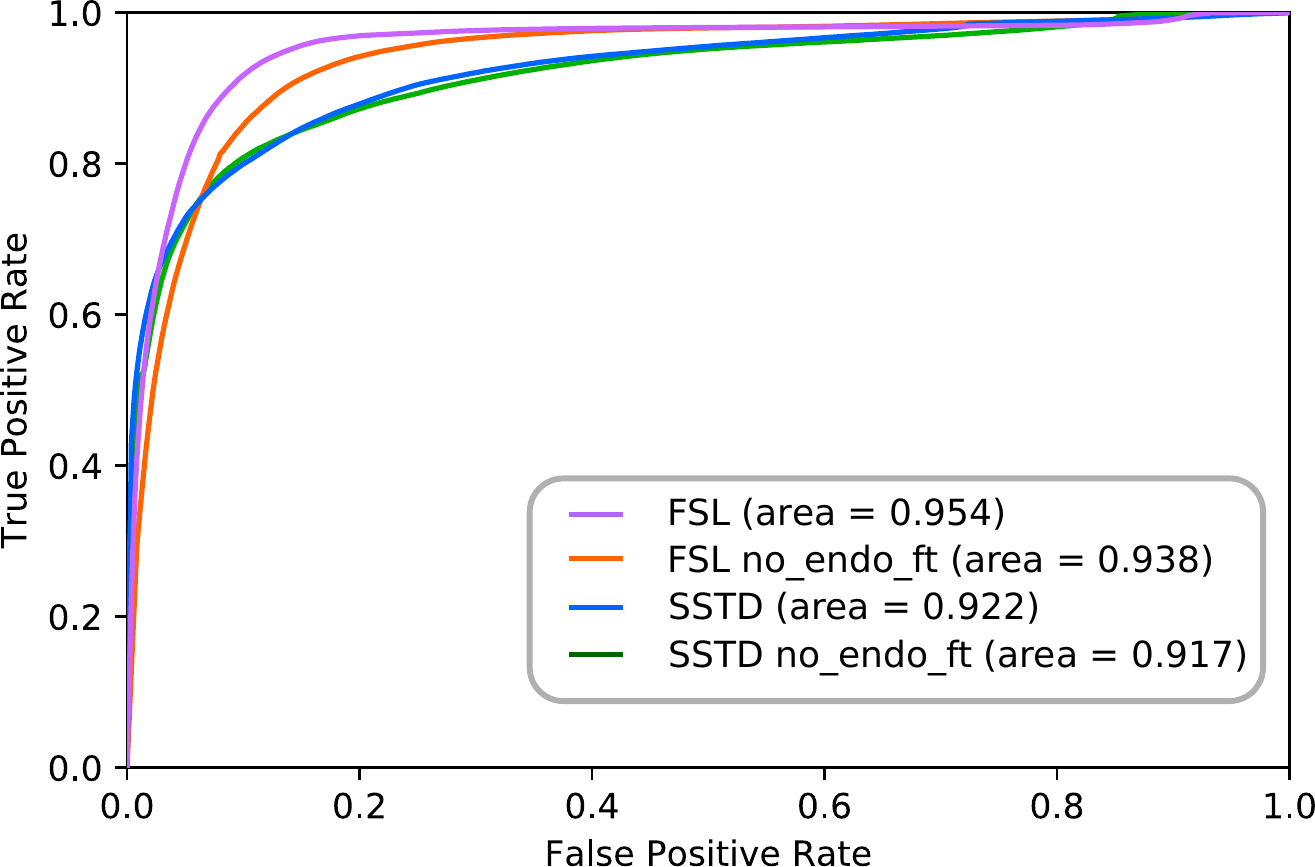}
  \caption{ROC curves for SSTS and FSL with and without pre-training on endoscopic data (no-endo-ft) on the in vivo dataset.}
  \label{fig:roc_curve}
\end{figure}





\begin{figure*}[!bht]  
  \centering
  \includegraphics[width=0.8\textwidth]{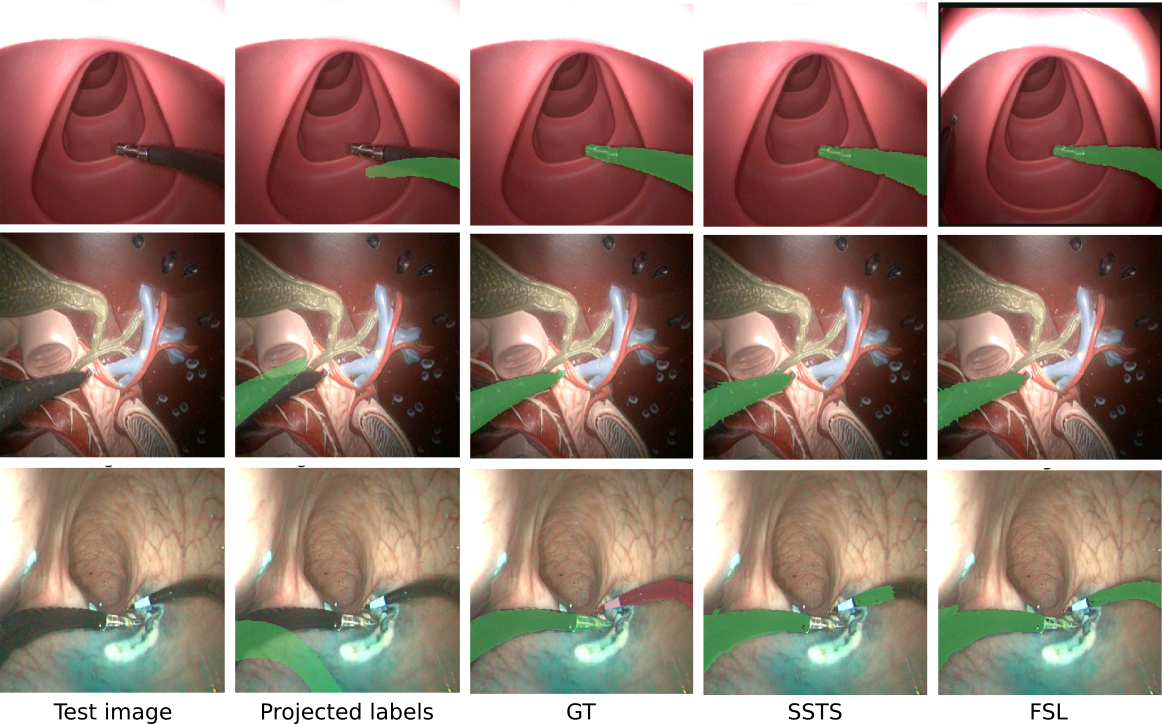}
  \caption{SSTS results on randomly selected images. Top: phantom 1 dataset; Middle: phantom 2; Bottom: in vivo. Best viewed in color.}
  \label{fig:results}
\end{figure*}

Table \ref{table:sstsresults} shows the comparison between the different approaches. For comparison, the output of the grabcut algorithm, initialized for each image from the projected labels obtained with the optimized transformation $T^*$, are also presented. As we hypothesized in section~\ref{sec:FCN}, even after optimizing the transformation $T^*$ the projected labels are not always sufficient to generate good results using the grabcut algorithm for inference. On the contrary, the proposed SSTS approach reached metrics which are similar to those of fully supervised learning (FSL) on the three datasets. One can note that the overall results obtained with both FSL and SSTS are lower on the \textit{in vivo} dataset. This is due to the fact that this datasets presents the most complex scenario, with specular reflections, smoke and body fluids. 

As detailed in section~\ref{sec:FCN}, the FCN structure is based on ResNet18, pre-trained on Imagenet and fine-tuned on endoscopic images using the approach from~\cite{Vardazaryan2018-du}. In order to assess the effect of this fine-tuning step on endoscopic images, we computed Receiver-Operating Characteristic (ROC) curves for the classification output, with and without using the endoscopic images for fine-tuning. Results are presented in Fig.~\ref{fig:roc_curve} for the in vivo dataset. One can see that the endoscopic fine-tuning slightly boosts performance with respect to the model trained on the Imagenet database. This highlights the benefits of seeing similar images during pre-training when the approach is applied on challenging in vivo data.

Figure~\ref{fig:results} shows example of results obtained on images from the three datasets. One can see that, in the event of contact with the tissues, the self-generated labels obtained with the kinematic model, even after optimization of $T$, can be quite far from the instrument. The result of the FCN training, however, is convincing and very close to the manually-annotated GT. It is worth noting that, on the \textit{in vivo} dataset, two robotic arms are present in the image. In this study only one arm was considered for training, however, due to a similar appearance, both SSTS and FSL segmented part of the right arm. The second instrument was therefore manually annotated using a different color (red overlay in Fig.~\ref{fig:results}) and such parts of the GT images were not considered for computing the metrics shown in Table~\ref{table:sstsresults}. We will extend the proposed algorithm for the bimanual case in future work.

\section{Conclusion and future work}

This paper presents a method for training a Fully Convolutional Neural Network in order to perform tool segmentation in robot-assisted surgery. Contrarily to most existing approaches, no manual intervention is required to do so. Instead, we make use of the approximate kinematic model of the robot to generate labels for training in a self-supervised manner. The proposed approach is done in three steps : first, a few images are acquired, together with the corresponding joint values of the robot, in the beginning of the surgery. The robot should be moving to allow for best efficiency. Second, a branch-and-bound optimization using a Grabcut-based cost function helps finding the hand-eye transformation $T^*$ which maximizes the IoU between the self-generated labels and the tool in the images. Finally, after convergence, such labels are used to fine-tune an FCN. The obtained classifier is then specific to the surgery being performed. Experimental evaluation was performed using a robotic flexible endoscope on two phantom and one in vivo dataset, showing very promising results, almost on par to fully supervised learning. 

While the results are promising, the proposed approach presents a few points which require further work. First, we will speed up the implementation for online use by investigating using a GPU implementation of the Grabcut algorithm, which promises a 10x speedup~\cite{Cheng2015-ah}, as well as by optimizing the architecture of the FCN. Using other information such as the gripping DOF of the robot, or the temporal dimension (e.g. relating motion features in the image with the robot displacement), might also enhance the results. Finally, we will investigate running SSTS at regular time points, in order to fine-tune the FCN as the surgery advances and cope with changing surgical environments. 



\bibliographystyle{IEEEtran}
\bibliography{test2}

\end{document}